\documentclass[letterpaper]{article} 
\usepackage{aaai23}  
\usepackage{times}  
\usepackage{helvet}  
\usepackage{courier}  
\usepackage[hyphens]{url}  
\usepackage{graphicx} 
\usepackage{adjustbox}
\usepackage{multirow}
\usepackage{natbib}  
\usepackage{caption} 
\usepackage{algorithm}
\usepackage{algorithmic}

\usepackage{newfloat}
\usepackage{listings}
\DeclareCaptionStyle{ruled}{labelfont=normalfont,labelsep=colon,strut=off} 
\floatstyle{ruled}
\newfloat{listing}{tb}{lst}{}
\floatname{listing}{Listing}
\title{Unified High-binding Watermark for Unconditional Image Generation Models}
\author{
    Ruinan Ma\textsuperscript{\rm 1},
    Yu-an Tan\textsuperscript{\rm 1},
    Shangbo Wu\textsuperscript{\rm 1},
    Tian Chen\textsuperscript{\rm 1},
    Yajie Wang\textsuperscript{\rm 1},
    Yuanzhang Li\textsuperscript{\rm 2}\thanks{Corresponding author.}
}
\affiliations{
    \textsuperscript{\rm 1}School of Cyberspace Science and Technology, Beijing Institute of Technology\\
    \textsuperscript{\rm 2}School of Computer Science and Technology, Beijing Institute of Technology\\
    $\left\{ruinan,tan2008,shangbo.wu,chentian20,wangyajie19,popular\right\}$@bit.edu.cn

}

\usepackage{bibentry}

\begin{document}

\maketitle

\begin{abstract}
Deep learning techniques have implemented many unconditional image generation (UIG) models, such as GAN, Diffusion model, etc. The extremely realistic images (also known as AI-Generated Content, AIGC for short) produced by these models bring urgent needs for intellectual property protection such as data traceability and copyright certification. An attacker can steal the output images of the target model and use them as part of the training data to train a private surrogate UIG model. The implementation mechanisms of UIG models are diverse and complex, and there is no unified and effective protection and verification method at present. To address these issues, we propose a two-stage unified watermark verification mechanism with high-binding effects for such models. In the first stage, we use an encoder to invisibly write the watermark image into the output images of the original AIGC tool, and reversely extract the watermark image through the corresponding decoder. In the second stage, we design the decoder fine-tuning process, and the fine-tuned decoder can make correct judgments on whether the suspicious model steals the original AIGC tool data. Experiments demonstrate our method can complete the verification work with almost zero false positive rate under the condition of only using the model output images. Moreover, the proposed method can achieve data steal verification across different types of UIG models, which further increases the practicality of the method. 
\end{abstract}

\section{Introduction}
In the past two years, AIGC~\cite{AIGC_survey_1,AIGC_survey_2} has become the hottest research direction in the field of deep learning, and has achieved impressive progress. AIGC tools represented by ChatGPT, Stable diffusion, and Gen-2 can assist users in creating highly creative and realistic images, videos, audios and other works of art with or without generating conditional information. The remarkable capabilities of AIGC tools benefit from the amount of training data and model size far exceeding traditional deep learning tasks. On the one hand, each AIGC tool owner needs to have the ability to trace the legal responsibilities and the source of the content generated by the tool; on the other hand, these high-quality generated contents are valuable intellectual property (IP) for the tool owner. At present, several AIGC tool publishers have paid attention to the potential data security issues of AIGC and released corresponding statements. For example, the legal provisions of ChatGPT clearly stipulate that it is strictly forbidden to use its output results as training data to train other models.\par

\begin{figure}[tbp]
\centering
\includegraphics[width=1.0\columnwidth]{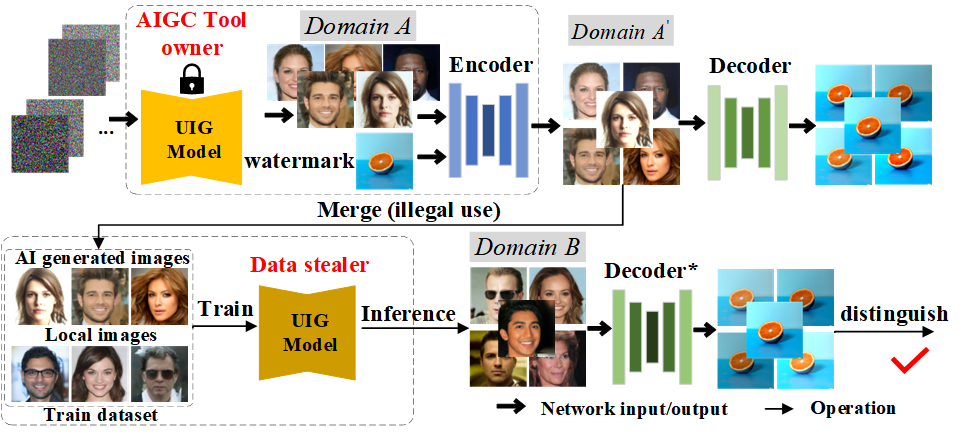}
\caption{The owner of the original unconditional image generation (UIG) model uses encoder and decoder to write and extract the watermark image from the output images. Once the watermarked images are misused, the fine-tuned decoder can be used to extract corresponding watermark image from the output images of the suspicious model.}
\label{pipeline_1}
\end{figure}

UIG models refer to a class of models that generate new samples which conform to the same distribution as the training data without conditional generation information, such as Denoising Diffusion Probabilistic Models (DDPM)~\cite{DDPM}, Generative Adversarial Network (GAN)~\cite{StyleGAN2} and Variational Autoencoder (VAE)~\cite{VQVAE2}. The diversity and authenticity of AI-generated images are the goals pursued by UIG models. State-of-the-art (SOTA) models can generate images so convincingly that it is difficult for humans to distinguish real images from AI-generated ones. Therefore, some AIGC tool manufacturers are very likely to use the output images of existing tools as training data to train their own model. Once such a thing happens, the interests of the original tool owner will be seriously violated. Effective supervision of such phenomena is a prerequisite for the healthy development of AIGC tools.\par

Recently, some research works starts paying attention to the IP protection of neural network. Frequently, they incorporate a regularization term into the loss function~\cite{watermark_network_1,watermark_network_2} or employ the forecasts of a specific group of indicator images\cite{GAN_BDR_watermarker,BDR_watermark_network} as the watermark. However, due to the uncertainty and diversity of AIGC, traceability and copyright protection of AIGC is still a seriously under-researched field. Different from image processing models~\cite{Image_process}, the input and output of UIG models belong to completely different image domains, usually using random noise as input to output high-quality images. The noise-image mapping functions formed by different UIG models under the same generation task may result in different output images for the same input noise. Therefore, such randomness makes previous model watermarking methods difficult to apply to current unconditional image generation tasks. Although some works have improved model watermarking for specific UIG models \cite{VAE_watermarker,GAN_BDR_watermarker,UIG_watermark}, these methods are not unified, that is, they are not applicable to all types of UIG models.\par

Motivated by this, we hope to protect the output images of UIG models, and in particular provide methods to protect the attribution and usage of such images. In a more general scenario, in order to expand the scale of training data so that the model can produce better image generation effects, data stealers may combine their existing image resources and the output images of the target AIGC tool to jointly obtain a private surrogate UIG model even package it as a commercial AIGC tool. In order to make the method more practical, the proposed watermark verification mechanism should be able to make a correct judgment on whether it has embezzled the data of the target AIGC tool only according to the output images of the suspicious AIGC tool. Furthermore, since it is not clear what type of models the data stealers use to implement their private AIGC tool, the proposed method should be applicable to different types of UIG models.\par

In this paper, we propose a unified high-binding watermark verification mechanism for UIG models to solve the AIGC data attribution problem existing in such models, as shown in Figure 1. Given a target model as the built-in model in the original AIGC tool, we denote its original output images distribution as domain $A$. In the first stage, also known as the preparation stage, the owner of the tool can use an encoder to embed a custom watermark image into the built-in model output images in a spatially invisible way. We define the distribution of the watermarked image as domain $A^{'}$. Assume that the data stealer combines his local images and the images collected from domain $A^{'}$ as total training dataset, and then uses it to train a surrogate UIG model with similar generation ability. We define the output image distribution of this surrogate UIG model as domain $B$. In the second stage, also known as the verification stage, when the original AIGC tool owner discovers that there is a suspicious model, he can verify whether there exists data steal based on whether the watermark image can be extracted from the output of the suspicious model. In particular, we use multiple losses in the first stage to embed the watermark more covertly into the image, and design a decoder fine-tuning strategy in the second stage to greatly improve the decoder's watermark extraction ability.\par

We conduct experiments using a variety of UIG models implemented by different mechanisms to demonstrate that our watermarking verification mechanism is applicable when different models are used as surrogate models. Experimental results show that our method can maintain a false positive rate close to 0\% while having a high watermark extraction rate, which shows that our method has very good practical usability.\par

Our contributions can be summarized as the following:
\begin{itemize} 
\item We are the first to introduce AIGC copyright protection problem for uncondational image generation tasks. We hope it can draw more attention to this research field and inspire greater works.
\item We propose a high-binding unified watermark verification mechanism for UIG models. The framework can accurately judge whether the suspicious UIG model abuses the original model output images only according to its output images.
\item Extensive experiments demonstrate our method is task-agnostic and plug-and-play for arbitrary UIG models. Furthermore, our method is capable of cross-model validation under different data theft situations.
\end{itemize}

\section{Related work}
\subsection{Image Watermarking Technology}
Watermarking technology represented by steganography is one of the most important ways to protect the copyright of images. It can be roughly categorized into two types: visible watermarks like logos, and invisible watermarks. Obviously, the way of invisible watermark is more covert than the way of adding visible watermark to images and does not affect the use of them. Invisible watermark can be further divided into transform domain watermark and spatial domain watermark. The transform domain watermark method writes watermark into the coefficients that are not significant in the transform domain such as Discrete Fourier Transform (DFT) domain~\cite{dft_stego} and Discrete Wavelet Transform (DWT) domain~\cite{dwt_stego} of the images. Spatial watermark~\cite{image_water1,swe_AIGC,IJCAI_Imagestego} mainly uses deep neural network to complete watermark embedding and extraction. The image watermarking method is easier to implement than the model watermarking method~\cite{BDR_watermark_network,watermark_network_1,watermark_network_3}, but it is not suitable for verifying whether the surrogate model has data steal.

\subsection{Unconditional Image Generation Models}
UIG models can create novel, original images that are not based on existing images without any specific input conditional information. As shown in Figure 2, this type of model learns the mapping functions from random noise to images through the training process, and then the inference stage uses sampling noise as a starting point to generate new images. This can be used in a variety of applications, such as improving image recognition algorithms\cite{dm_use1} or generating realistic images to augment dataset. Variational Autoencoder (VAE)~\cite{VAE,VQVAE2} and Generative Adversarial Network (GAN)~\cite{GAN_ori,StyleGAN2} are two typical UIG models that appeared earlier, they realize the conversion between noise and images in a single step during training and inference. Diffusion models~\cite{DDPM,DDIM,PNDM} draw inspiration from the principles of non-equilibrium thermodynamics. They establish a Markov chain of diffusion steps that gradually introduce random noise into image, and subsequently acquire the ability to invert the diffusion process to generate desired image samples from the noise. Although these UIG models achieve the same purpose, their different and complex implementation mechanisms bring difficulties to the formation of an unified watermark verification mechanism.

\subsection{Watermark for AIGC}
Though AIGC Watermarking Technology is still seriously under-studied, there are some initiatives and works that start paying attention to it. The famous AIGC tool DALLE marks the generated content by adding a spatially visible watermark in the lower right corner of the image. Stable Diffusion modifies specific Fourier frequency in the generated image~\cite{SB_watermark} to watermark output images. \cite{dnn_stego2} proposes an universal adversarial watermark to protect images from being abused by Deepfake technology. \cite{swe_AIGC} proposes a non-embedded watermark method, so that the newly generated image naturally contains potential watermark information. Existing work mainly focuses on the direct traceability of AIGC, and very few works~\cite{train_w1, train_w2} study how to effectively verify AIGC data after it has been abused. At present, there is no unified watermark verification mechanism that can realize AIGC copyright protection across UIG models.


\begin{figure}[htbp]
\centering
\includegraphics[width=1.0\columnwidth]{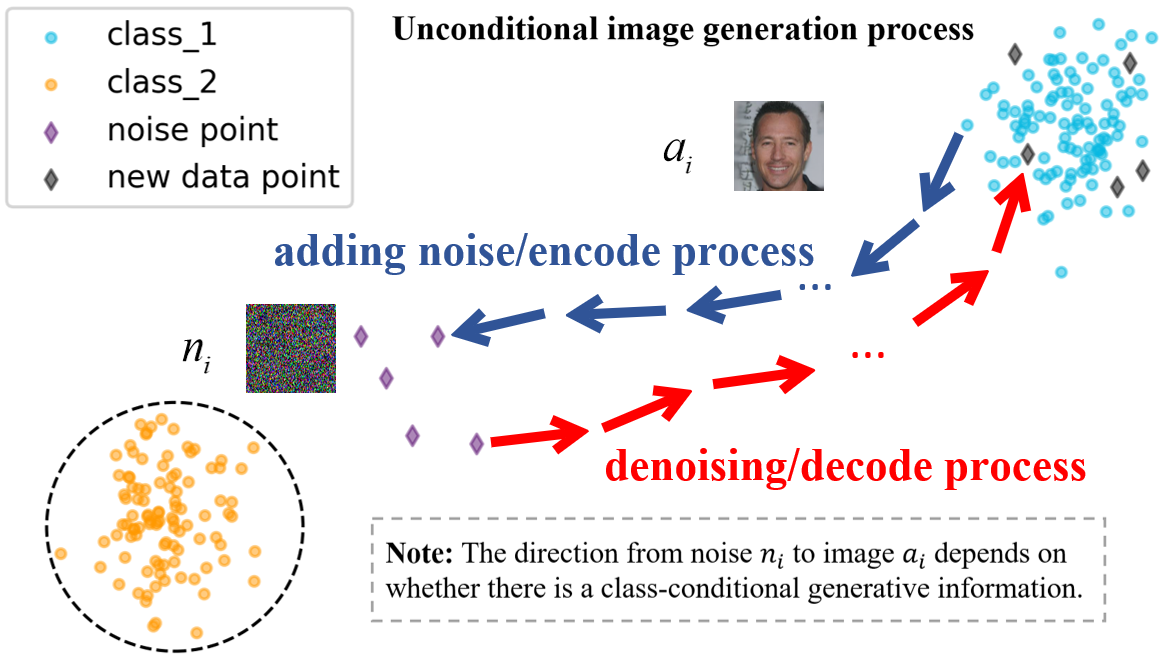}
\caption{The training process of the UIG model can be regarded as an encoding process from existing training images to random noise. The inference process of the trained UIG model can be seen as the process of decoding from randomly sampled noise to new images.The number of encoding and decoding steps depends on the UIG model mechanism.}
\label{diffusion_process_1}
\end{figure}

\begin{figure*}[htbp]
\centering
\includegraphics[width=17cm, height=9cm]{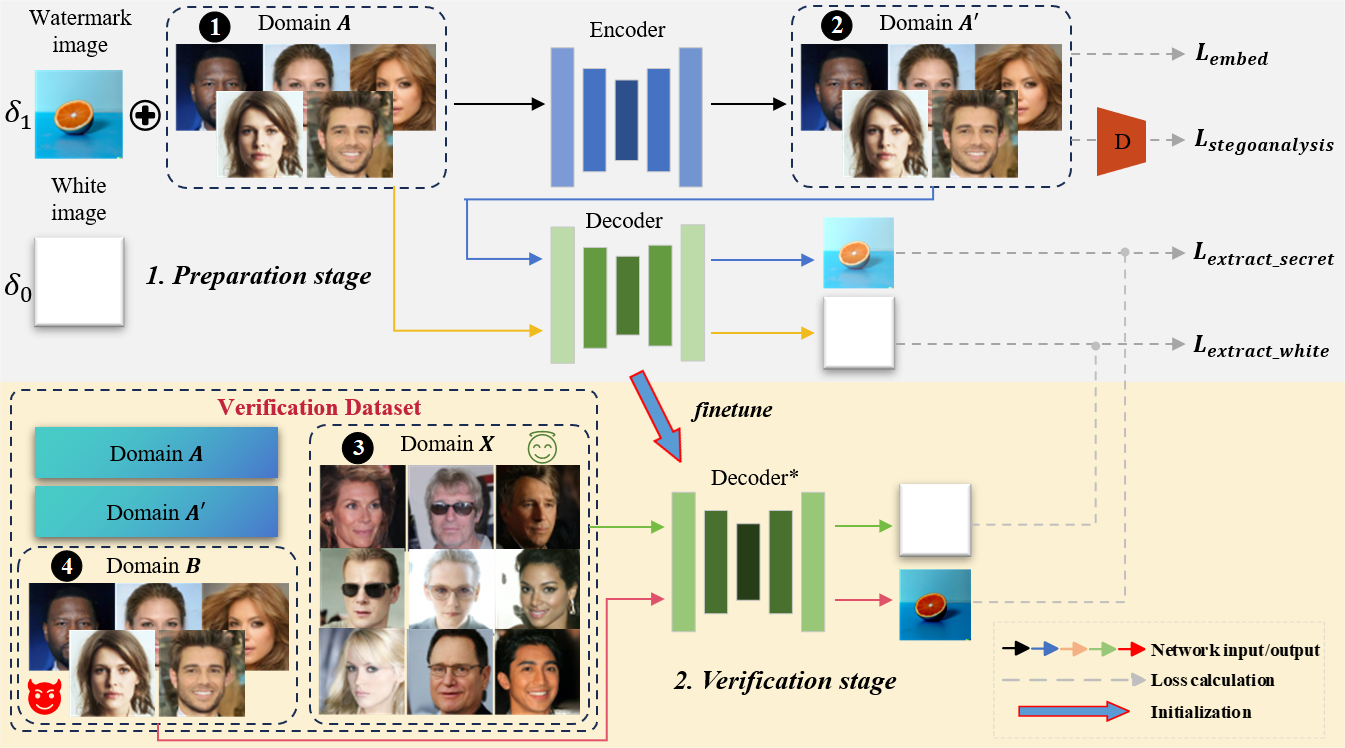}
\caption{The overall pipeline of the proposed two-stage unified watermark verification mechanism with high-binding effects for unconditional image generation models.The fine-tuned decoder can make correct judgments on whether the suspicious model steals the original AIGC tool data only depends on the model output images.}
\label{method}
\end{figure*}

\section{Method}
In this section, we first introduce a formal problem definition, and then elaborate on the roles and implementation details of each stage of the proposed unified two-stage watermark verification mechanism.

\subsection{Problem Definition} For an UIG model, we denote its output images as domain $A$ according to the image distribution, and domain $A$ follows the same distribution as the model training data. The goal of the UIG model $M$ is to approximately fit the mapping function $F$ from random noise to the target image distribution through one or several neural networks. Suppose we obtain an UIG model $M$ trained on a private dataset $Trainset$. Given a random noise $n_{i}$, $M$ can output an image $a_{i}$ in domain $A$ that has never actually appeared in $Trainset$, i.e.,
\begin{equation}
    M(n_{i})=a_{i}\sim domain\;A, a_{i}\notin Trainset
\end{equation}\par
Since UIG models use random noise as the starting point to generate new images, the attacker only needs to collect the output images of the target model instead of the input-output image pairs to train the corresponding surrogate model $SM$, which greatly reduces the difficulty for the attacker to steal. In addition, according to the image distribution, the output images of $SM$ belong to another image domain with similar but different distribution to domain $A$. In practical scenarios, AIGC tools formed from such models are usually presented in a black-box manner, and users can only obtain their output images. Therefore, our goal is to design a unified watermark verification mechanism with high-binding effect across data domains and even across model architectures, which can verify whether it steals $M$ generated images for training only through the outputs of the suspicious model.

\subsection{Watermark Embedding Process}
This is the first stage in our proposed unified watermark verification mechanism, which can also be called the preparation stage. For data stealers, they can only get the output images of the target AIGC tool instead of the built-in UIG model. For the suspicious UIG model that we mentioned above, the publisher of the original AIGC tool can only verify whether there is data steal based on its output images. As can be seen, the model output images are the only medium between the original AIGC tool owner and the suspect model. Therefore, we hope that the watermark image used as the basis for verification can be completely embedded in the output images of the above two types of models. In particular, the watermark image should be embedded in an invisible way so as not to affect the use of the model output image.\par

In the preparation stage, we first ensure that the output images of the original UIG model is embedded with a custom watermark image $\delta _{1}$. As shown in the upper part of Figure 3, we select a neural network as an encoder $H$ to embed $\delta _{1}$ into the output images of the protected UIG model (i.e., domain $A$) in spatial invisible way, and denote the distribution of watermarked images as domain $A^{'}$. Obviously, the distributions of domain $A$ and domain $A^{'}$ are similar, the subtle difference between the two distributions is caused by the potential watermark information. At the same time, we choose another neural network as the decoder $R$ to reversely extract $\delta _{1}$ from images in domain $A^{'}$.\par

The difference between the corresponding image pair in domain $A$ and domain $A^{'}$ is extremely small (i.e., spatial invisible), the watermark image $\delta _{1}$ is injected into the images of domain $A$ mainly by changing the high-frequency information. In order to improve the ability of $R$ to recover the specified watermark image $\delta _{1}$ and to avoid the overfitting phenomenon that $\delta _{1}$ can be extracted from all images no matter whether $\delta _{1}$ is really written or not, we simultaneously add images from domain $A$ to the dataset for training $R$. We force $R$ to extract a constant blank images $\delta _{0}$ from these non-watermarked images. To further narrow the data distribution difference between domain $A$ and domain $A^{'}$, we add a steganalysis network $D$ as a discriminator after $R$, and improve image quality in domain $A^{'}$ through adversarial training.\par

After the preparation stage, the publisher of the original AIGC tool can use $H$ to embed $\delta _{1}$ in the original output image of the model and then output it. In this way, what the data stealers get is the output images with the watermark information. The corresponding $R$ can be used to directly trace the source of the released images.

\subsubsection{Network Structures.}
In our method, for the encoder $H$, the input and output image domains are highly similar, so we choose U2-Net~\cite{U2NET} as the default network structure. For the decoder $R$, we choose CEILNet~\cite{CEILNet} as the network structure to cope with image translation across image domains. As for the discriminator $D$, we choose the steganalysis network SRNet~\cite{SRNet} to ensure that the concealment of watermark embedding is increased while improving the image quality. To verify the generalization ability of the proposed method across UIG models, we choose UIG models with the same or different model structure as the original AIGC tool built-in model to obtain $SM$.

\subsubsection{Loss Functions.}
The objective loss function of watermark embedding stage consists of two parts: the embedding loss $l_{emb}$ and the extracting loss $l_{ext}$, i.e.,
\begin{equation}
    l=l_{emb}+l_{ext},
\end{equation}
The purpose of embedding loss is to ensure that the visual quality of the image is not affected after the watermark image is embedded into the output images of the original UIG model, two different types of visual consistency loss are considered: the basic embed loss $l_{embed}$ and steganalysis adversarial loss $l_{sadv}$, i.e.,
\begin{equation}
    l_{emb}=\lambda_{1}*l_{embed}+\lambda_{2}*l_{sadv},
\end{equation}
where $\lambda_{1}$ and $\lambda_{2}$ are hyperparameters to balance these two loss terms. $l_{embed}$ is used to calculate Mean-Squared Loss (MSE) for the image $a_{i}$ from domain $A$ and the corresponding image $a_{i}^{'}$ from domain $A^{'}$, i.e.,
\begin{equation}
    l_{embed}=\sum\limits_{a_{i}\in A, a_{i}^{'}\in A^{'}}MSE(a_{i}, a_{i}^{'}),
\end{equation}
$l_{sadv}$ will let discriminator $D$ cannot differentiate the output images of $H$ from the images in domain $A$, which makes the watermark embedding process more covert.
\begin{equation}
    l_{sadv}=\mathbf{E}_{a_{i}\in A}log(D(a_{i}))+\mathbf{E}_{a_{i}^{'}\in A^{'}}log(1-D(a_{i}^{'})).
\end{equation}
The purpose of the extracting loss is to enable $R$ to reversely extract the specified watermark image $\delta _{1}$ from the output images of $H$ and extract the constant blank image $\delta _{0}$ from the original image without watermark embedded, i.e.,\\
\begin{equation}
    l_{ext}=\sum\limits_{a_{i}\in A}MSE(D(a_{i}),\delta _{0}) +\sum\limits_{a_{i}^{'}\in A^{'}}MSE(D(a_{i}^{'}),\delta _{1}).
\end{equation}

\subsection{Decoder Fine-tuning Strategy}
This is the second stage of our proposed unified watermark verification mechanism, which can also be called the verification stage. After this stage, we hope that the fine-tuned $R$ can also extract $\delta _{1}$ from the output images of $SM$.\par
As shown in the lower part of Figure 3, as the verifier, the publisher of the original AIGC tool needs to prepare a verification dataset containing four parts of the images, i.e., images from domain $A$, $A^{'}$, $B$, and $X$ respectively. According to the above, the images in domain $A$ and domain $A^{'}$ are naturally exist after the preparation stage, especially the images of domain $A$ are private to the AIGC tool owner. The images in domain $B$ are collected from the output of the suspicious model to be verified. The images of domain $X$ come from the same task UIG models (preferably with different model architectures), which are trained on images without embedded watermarked images. It should be noted that these UIG models for the same task can be easily prepared by the AIGC tool owner based on the images of domain $A$, so they are also private.\par

After the validation dataset is ready, we finetune the decoder $R$ using the decoder fine-tuning strategy. It consists of two parts the watermark extraction loss $l_{e\_s}$ and the constant blank image extraction loss $l_{e\_w}$, i.e.,
\begin{equation}
    l_{e\_s}=\sum\limits_{a_{i}^{'}\in A^{'}}MSE(D(a_{i}^{'}),\delta _{1})+\sum\limits_{b_{i}\in B}MSE(D(b_{i}),\delta _{1}),
\end{equation}
\begin{equation}
    l_{e\_w}=\sum\limits_{a_{i}\in A}MSE(D(a_{i}),\delta _{0}) +\sum\limits_{x_{i}\in X}MSE(D(x_{i}),\delta _{0}).
\end{equation}

\subsubsection{Data Theft Verification.} With the above two-stage training process, we obtain the fine-tuned decoder $R^{'}$. The verifier uses $R^{'}$ to extract the watermark image from the output images of the target $SM$, if the similarity between the output image of $R^{'}$ and the original real watermark image $\delta _{1}$ is greater than the preset threshold, it can be determined that the $SM$ has data steal.

\section{Experiment}
In this section, we will frst describe our datasets and implementation details. Following, we introduce our evaluation metrics. Then, we show the experimental results of the proposed two-stage unified watermark verification mechanism for UIG models. Moreover, we systemically conduct an ablation study.

\subsection{Datasets and Implementation Details}
In our experiments, we use CelebA~\cite{CelebA} trainset as the main dataset, which is more challenging, has higher resolution and contains more texture details in the images. It contains 202599 facial images. These images are split into two parts: 50,000 images are owned by the data stealer, others are used to train the original UIG models. All images in CelebA are processed into 128*128. In addition, we also verified the effectiveness of the method in another Oxford 102 Flower dataset~\cite{flower102}. It contains 8189 real flower images, we set 2000 as private images for data stealer, others are used to train original UIG models. All images in Oxford 102 Flower are processed into 64*64.\par

The original UIG models we select in our experiments are VQ-VAE-2~\cite{VQVAE2}, StyleGAN2~\cite{StyleGAN2}, DDPM~\cite{DDPM} and PNDM~\cite{PNDM}. The selected models are typical models of various UIG models under different implementation mechanisms. For the diffusion model that generates new samples through multiple steps, We choose DDPM and PNDM to consider the impact that different noise sampling strategies may have on the performance of the proposed method. \par

For the validation dataset involved in the second stage of our method, we always keep each part of the dataset to contain 1000 images when using the CeleA dataset, and we always keep each part of the dataset to contain 100 images when using the Oxford 102 Flower dataset. In all experiments, we fine-tune the decoder using the validation dataset until convergence is confirmed. We use new images outside the validation dataset to evaluate watermark extraction. By default, $\lambda_{1}$ and $\lambda_{2}$ equal to 1. 

\subsection{Evaluation Metrics}
To evaluate the visual quality, PSNR and SSIM are used by default. In order to judge whether the watermark image is extracted from the image, we define that the extraction is successful when the SSIM between the extraction result and the real watermark image is bigger than 0.9, which means that there is a large similarity between them. Based on it, the extraction rate (ER) is further defined as the ratio of successfully extracting watermark image from model output images. The false positive rate (FPR) is defined as the ratio of successfully extracting watermark image from model output images without data steal. Furthermore, we define the data theft rate (DTR) as the ratio of output images from the original AIGC tool in the surrogate model training data.

\subsection{The Results of Our Method}
\subsubsection{Invisible Watermark Embedding.} This experiment is to demonstrate the effect of watermark embedding in the first stage of our method. In order to prove that our method has the ability to embed large-capacity watermark image, we choose three-channel color images with more information as the watermark image (for single-channel grayscale images, our method is still applicable). We use the trained encoder $H$ to invisibly embed the watermark image into the output images of the UIG model, and then use the trained decoder $R$ to inversely extract the watermark image, two visual examples are shown in Figure 4. For the quantitative results, we randomly selected 1000 original face generation UIG model output images for testing (because of the high resolution). The average SSIM between the watermarked image and the original image is 0.95, and the average PSNR is 38.66. The average SSIM of the extracted watermark image and the original watermark image is 0.98, and the average PSNR is 42.32. This shows that through our watermark embedding process, it does not affect the use of the original images.
\begin{figure}[htbp]
\centering
\includegraphics[width=1.0\columnwidth]{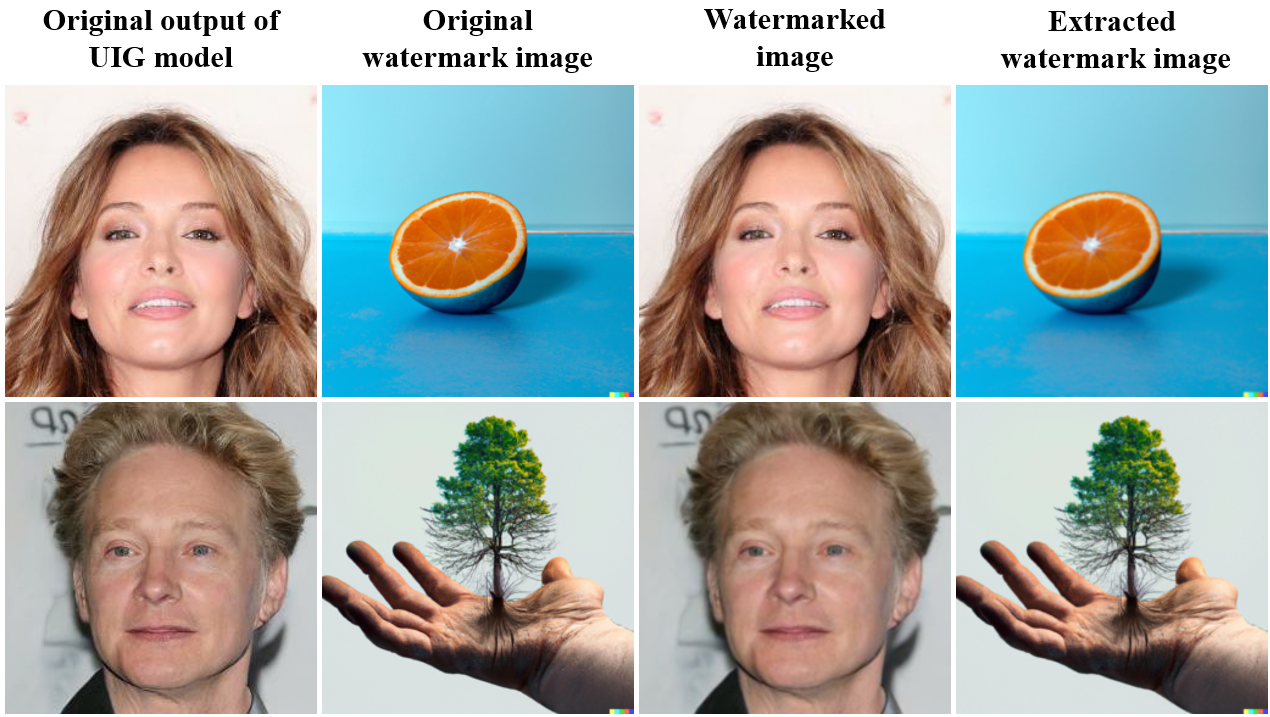}
\caption{Example of encoder $H$ and decoder $R$ input and output visualization results after the first stage. Watermark image is embedded invisibly and can be extracted perfectly.}
\label{figure4}
\end{figure}

\subsubsection{Robustness to Different Structures of Surrogate Models.}
Since the data stealers may not know what model structure the original AIGC tool built-in model adopts, after collecting its output images, the data stealers may use other different network structures to build their private UIG model. To demonstrate the cross-model capability of our watermark verification mechanism, i.e. its generalizability to different structure UIG models, we employ many surrogate models to imitate real-world scenarios, and assume each model has partially private training data. \par

First, we choose a model as the original face generation UIG model and collect the images generated by it. Then, we jointly combine the collected images and private images into a training dataset for the surrogate model according to different DTRs. Next, we use these dataset to train different surrogate models and collect the output images of the surrogate models, which are used to form the validation dataset under the second stage of our method. After fine-tuning the decoder using the validation dataset, we perform ER calculations on the corresponding surrogate models using $R^{'}$. As shown in Figure 5, although ER decreases with the decrease of DTR, the $R^{'}$ obtained by our proposed method always maintains a high ER at any time, even when the DTR is only 10\%, the ER still exceeds 30\%. It is noteworthy that our method works better when DTR is larger. In practice, data stealers usually steal the images generated by AIGC tools because real images are difficult to collect. Therefore, the phenomenon of data steal in reality usually corresponds to the situation when the DTR is large in our experiment.\par

Experimental results show that our method has good cross-model capability, and it is a unified watermark verification mechanism for unconditional image generation models. The inherent multi-step iterative mechanism of the diffusion model makes it more difficult to learn the potential watermark features in the image. Therefore, with the decrease of DTR, the corresponding ER attenuation is more obvious.

\begin{figure}[htbp]
\centering
\includegraphics[width=1.0\columnwidth]{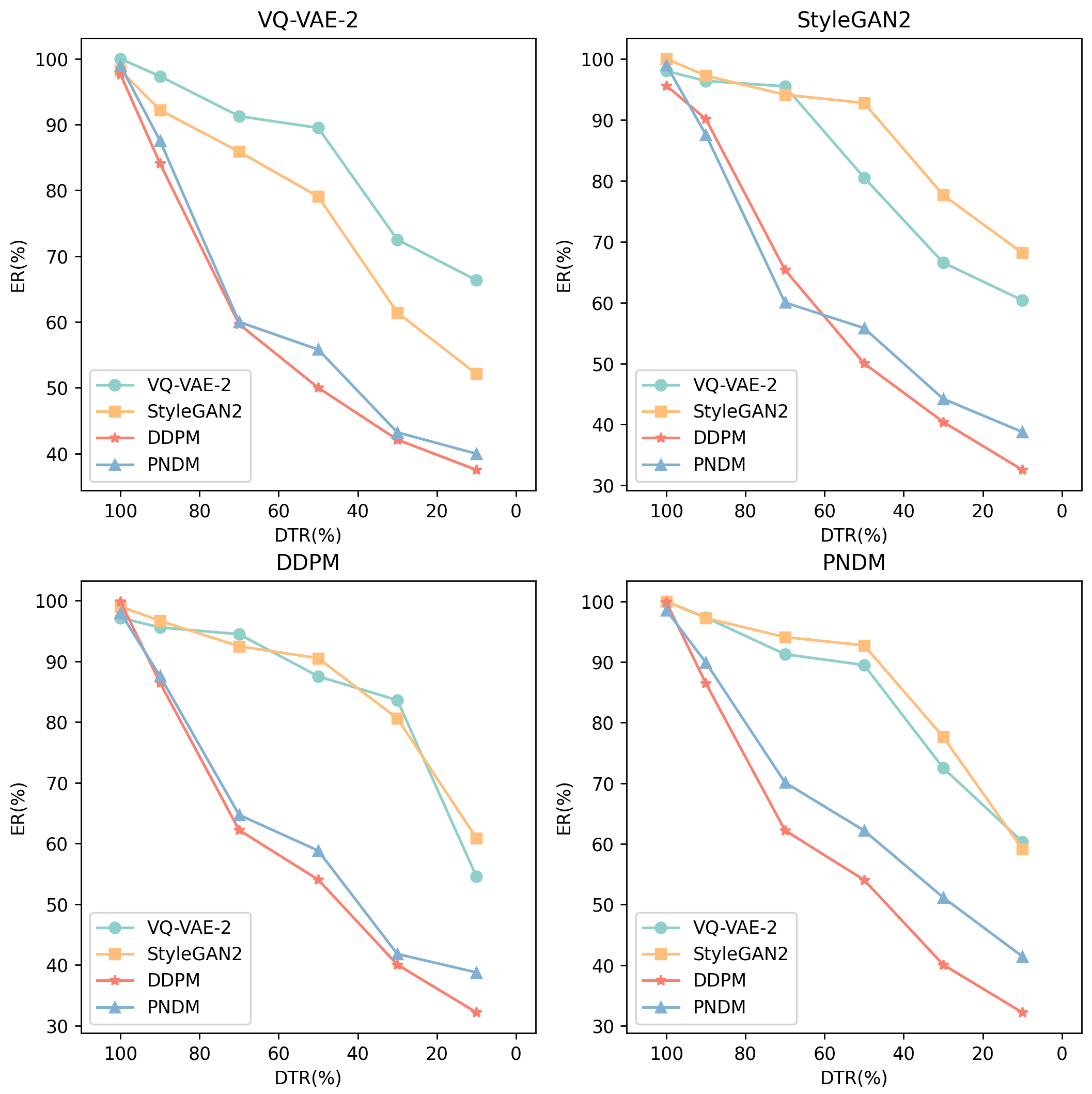}
\caption{ER curves of the decoder $R^{'}$ for face generation surrogate models with different structures under different DTR. We use 200 new images of the surrogate model each time we calculate the ER value.}
\label{cross_model}
\end{figure}

\subsection{Ablation Study}
\subsubsection{The Importance of Decoder Fine-tuning Stratege.} In order to improve the ability of the decoder $R$ to extract the target watermark image from the output images of the surrogate model, we use a fine-tuning strategy on decoder $R$ to obtain $R^{'}$. To demonstrate its necessity, we conducted experiments when the surrogate model used the same structure model as the original UIG model. With and without decoder fine-tuning (i.e., the second stage), we tested the watermark ER of the decoder on corresponding surrogate model under different DTRs, and the results are shown in Table 1. It can be seen that with the DTR gradually decreases, the ER of $R$ on the corresponding surrogate model gradually decreases. When the DTR is 10\%, $R$ completely loses its verification ability. However, the fine-tuned $R^{'}$ can recover the extraction ability substantially under various DTRs. We also train surrogate models on unwatermarked images, which we call $clean$ $SM$. It should be noted that in this experiment, the FPR of all decoders before and after fine-tuning for $clean$ $SM$ is 0\%. The visualization of the decoder extraction results before and after fine-tuning is shown in Figure 6. The important role of the decoder fine-tuning strategy is clear at a glance.

\begin{table}[htbp]
\begin{adjustbox}{width=\linewidth}
\begin{tabular}{c|cc|cc|cc}
\hline
\multirow{2}{*}{ER} & \multicolumn{2}{c|}{100\%} & \multicolumn{2}{c|}{50\%} & \multicolumn{2}{c}{10\%} \\ \cline{2-7} 
 & \multicolumn{1}{c|}{Stage1} & Stage2 & \multicolumn{1}{c|}{Stage1} & Stage2 & \multicolumn{1}{c|}{Stage1} & Stage2 \\ \hline
VQ-VAE-2 & 72.65\% & 100.00\% & 30.66\% & 89.50\% & 0.00\% & 66.36\% \\ \hline
StyleGAN2 & 60.33\% & 100.00\% & 10.26\% & 92.75\% & 0.00\% & 68.15\% \\ \hline
DDPM & 49.12\% & 99.87\% & 18.39\% & 54.05\% & 0.66\% & 32.18\% \\ \hline
PNDM & 56.46\% & 98.55\% & 22.92\% & 62.19\% & 1.45\% & 41.43\% \\ \hline
\end{tabular}
\end{adjustbox}
\caption{Watermark extraction rate (ER) of face generation surrogate models before and after decoder fine-tuning under different data theft rates (DTR). Stage1 means using $R$ as the decoder, Stage2 means using $R^{'}$ as the decoder.}
\label{tab:table1}
\end{table}

\begin{figure}[htbp]
\centering
\includegraphics[width=1.0\columnwidth]{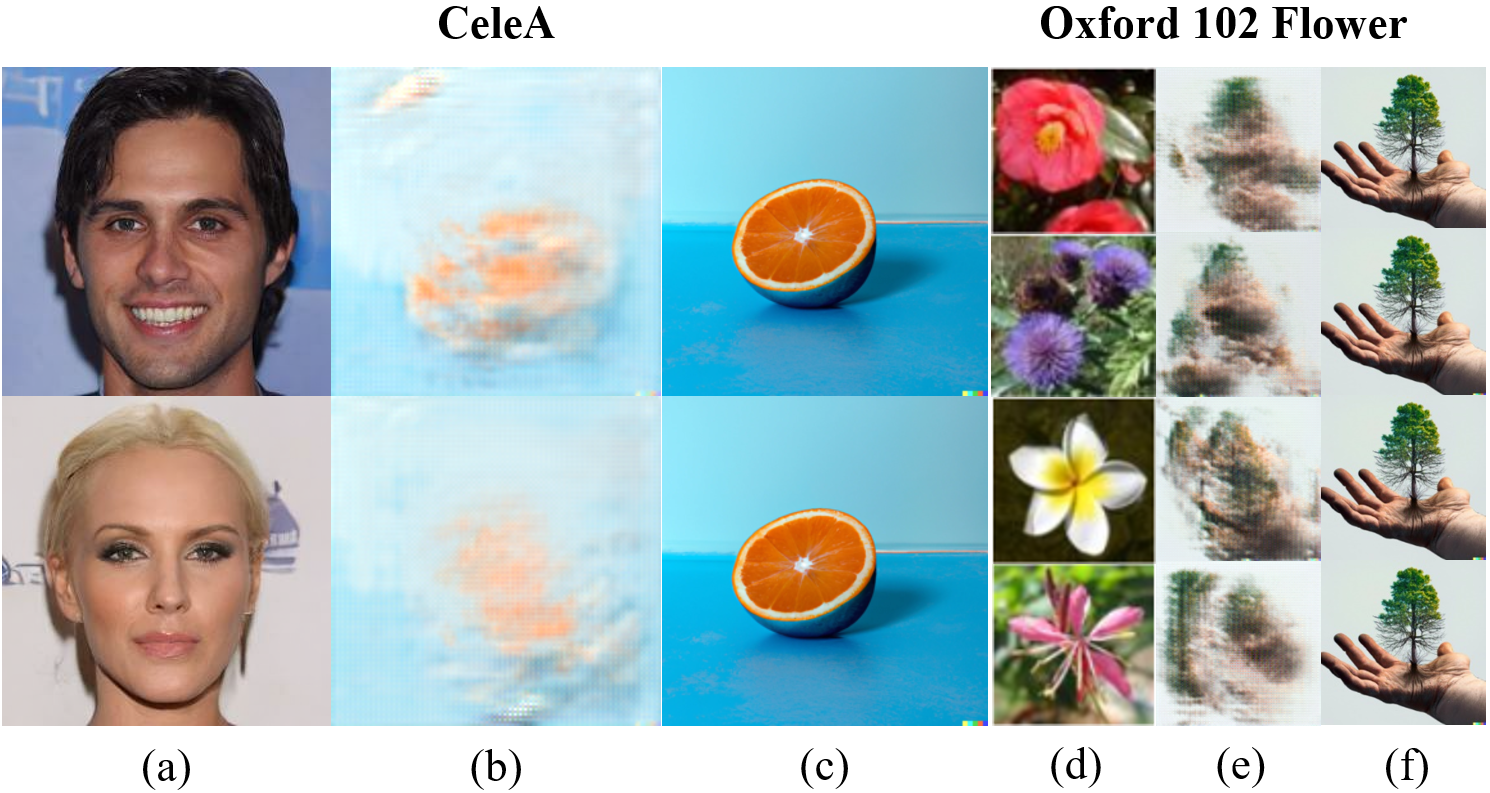}
\caption{The visualization of the decoder extraction results before and after fine-tuning.(a)(d)Original output images of UIG model. (b)(e)The extraction results of the decoder for the output image of the surrogate model before fine-tuning. (c)(f)The extraction results of the decoder for the output image of the surrogate model after fine-tuning.}
\label{figure5}
\end{figure}

\subsubsection{The Importance of Steganalysis Adversarial Loss.}
In order to make the embedding of watermark information more concealed, we added a steganalysis adversarial loss $l_{sadv}$. The most intuitive benefit it brings is the improvement of the visual quality of the image after embedding the watermark image. Form CeleA, the average SSIM and PSNR between the original model output image and the watermarked image increased from 0.94, 36.68 to 0.95, 38.66, respectively. For Oxford 102 Flower, the corresponding metrics improved from 0.92, 33.15 to 0.96, 39.72.\par

Furthermore, we visualized the difference between the watermarked image obtained with and without $l_{sadv}$ and the original image. As shown in Figure 7, after using $l_{sadv}$, the embedded watermark information is more focused on changing the pixels at key positions in the image, such as the face detail area in the image. We then analyze why our proposed method has high-binding effect. For UIG models, taking the face generation task as an example, the spatial face features in the training data are the main focus of the model's learning. Therefore, we concentrate the watermark information on key positions in the image, which is the key reason why the watermark image can be extracted from the output images of the surrogate model. This is also the reason why cross-model validation can be achieved.

\begin{figure}[htbp]
\centering
\includegraphics[width=1.0\columnwidth]{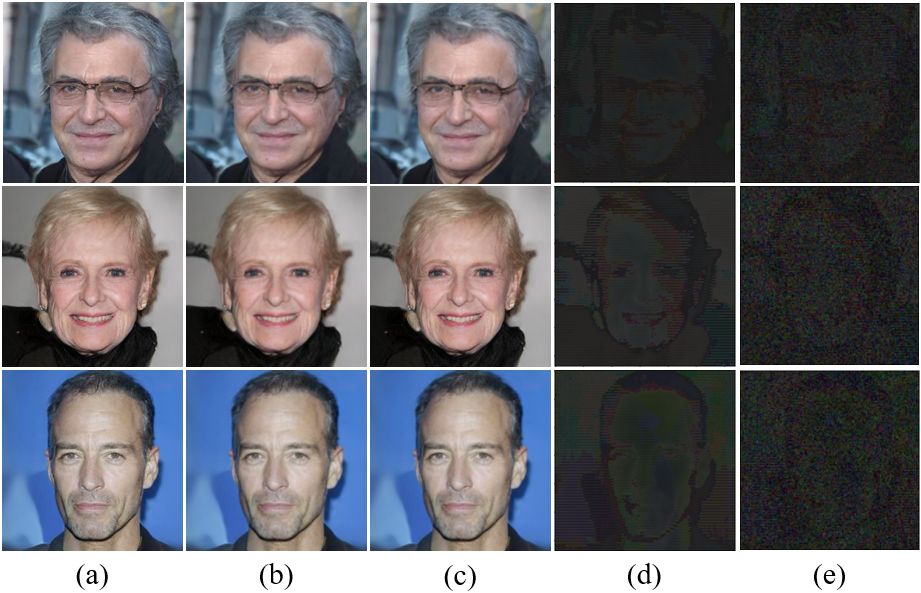}
\caption{Visualization of the difference between images before and after watermark embedding.(a)Original output images of UIG model. (b)(c)Images after embedding the watermark(with/without $l_{sadv}$). (d)The difference between (a) and (b). (e)The difference between (a) and (c). Note: All differences are magnified 20x for easier visualization.}
\label{difference}
\end{figure}

\subsubsection{The Importance of Constant Blank Image Extraction Loss.}
In order to reduce the overfitting phenomenon in the process of decoder fine-tuning. We add $l_{e\_w}$ to the decoder fine-tuning strategy to reduce the FPR of $R^{'}$ on surrogate models. We conduct experiments when the surrogate model has the same structure as the original model and the DTR is equal to 70\%, the results are shown in Table 2. We use a total of 400 images from domain $A$ and domain $X$ that did not appear in the validation dataset to count FPR. It can be seen that under different circumstances, with the help of $l_{e\_w}$, the FPR of $R^{'}$ is reduced to almost 0\%. 

\begin{table}[htbp]
\begin{adjustbox}{width=\linewidth}
\begin{tabular}{c|c|c|c|c}
\hline
FPR & VQ-VAE-2 & StyleGAN2 & DDPM & PNDM \\ \hline
\multirow{2}{*}{CeleA} & 14.25\% & 12.75\% & 8.25\% & 9.00\% \\
 & 0.00\% & 0.5\% & 0.00\% & 0.25\% \\ \hline
\multirow{2}{*}{\begin{tabular}[c]{@{}c@{}}Oxford 102\\ Flower\end{tabular}} & 22.50\% & 18.00\% & 11.25\% & 17.50\% \\
 & 0.25\% & 0.00\% & 0.00\% & 0.00\% \\ \hline
\end{tabular}
\end{adjustbox}
\caption{The false positive rate(FPR) of $R^{'}$ with and without $l_{e\_w}$ when the surrogate model has the same structure as the original model and DTR is equal to 70\%.}
\end{table}

\section{Conclusion}
In this work, we propose a high-binding unified watermark verification framework for unconditional image generation models. This is the first time that the AIGC copyright protection problem is introduced for the task of unconditional image generation. We use a two-stage process including spatially invisible watermark embedding and decoder fine-tuning to effectively address the verification of data steal. Extensive experiments prove that our method is task-agnostic and plug-and-play for arbitrary unconditional image generative models. We hope our work can draw more attention to this research field and inspire greater works.

\bibliography{aaai23}

\end{document}